\documentclass{article}

\usepackage{arxiv}

\usepackage[utf8]{inputenc} % allow utf-8 input
\usepackage[T1]{fontenc}    % use 8-bit T1 fonts
\usepackage[hidelinks]{hyperref}
\usepackage{url}            % simple URL typesetting
\usepackage{booktabs}       % professional-quality tables
\usepackage{amsfonts}       % blackboard math symbols
\usepackage{nicefrac}       % compact symbols for 1/2, etc.
\usepackage{microtype}      % microtypography
\usepackage{graphicx}
\usepackage{natbib}
\usepackage{doi}

\usepackage{tikz}
\usepackage{todonotes}
\usepackage{csquotes}
\usepackage{amsmath,amssymb,amsmath}
\usepackage{isomath}
\usepackage{pifont}
\usepackage{fontawesome}
\usepackage[noabbrev,nameinlink]{cleveref}

\def\etal{\emph{et al}.}

\newcommand{\cmark}{\ding{51}}
\newcommand{\xmark}{\ding{55}}

\newcommand{\orcid}[1]{\href{https://orcid.org/#1}{\includegraphics[scale=0.06]{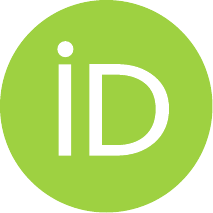}}}

\title{Higher-Order Adversarial Patches for Real-Time Object Detectors}

\author{
	Jens Bayer$^{1,2 \text{ \orcid{0000-0002-2806-6920}
	\href{mailto:jens.bayer@iosb.fraunhofer.de}{\tiny\faEnvelope}}}$
	\And
Stefan Becker$^{1 \text{ \orcid{0000-0001-7367-2519}}}$\And
David M\"unch$^{1 \text{ \orcid{0000-0002-8577-5256}}}$\And
Michael Arens$^{1 \text{ \orcid{0000-0002-7857-0332}}}$\And
J\"urgen Beyerer$^{1,2 \text{ \orcid{0000-0003-3556-7181}}}$\And
	\\
    $^1$ Fraunhofer IOSB and Fraunhofer Center for Machine Learning,\\
    $^2$ Karlsruhe Institute of Technology
}

\renewcommand{\shorttitle}{Higher-Order adv. Patches for Real-Time Object Detectors}

\hypersetup{
pdftitle={Higher-Order Adversarial Patches for Real-Time Object Detectors},
pdfsubject={},
pdfauthor={Jens Bayer, Stefan Becker, David M\"unch, Michael Arens, J\"urgen Beyerer},
pdfkeywords={adversarial attacks, adversarial training, object detection},
}

\begin{document}
\maketitle

\begin{abstract}
Higher-order adversarial attacks can directly be considered the result of a
cat-and-mouse game -- an elaborate action involving constant pursuit, near
captures, and repeated escapes. This idiom describes the enduring circular
training of adversarial attack patterns and adversarial training the best. The
following work investigates the impact of higher-order adversarial attacks on
object detectors by successively training attack patterns and hardening object
detectors with adversarial training. The YOLOv10 object detector is chosen as a
representative, and adversarial patches are used in an evasion attack manner.
Our results indicate that higher-order adversarial patches are not only
affecting the object detector directly trained on but rather provide a
stronger generalization capacity compared to lower-order adversarial patches.
Moreover, the results highlight that solely adversarial training is not
sufficient to harden an object detector efficiently against this kind of
adversarial attack.

Code: \url{https://github.com/JensBayer/HigherOrder}

\end{abstract}

% keywords can be removed

\section{Introduction}
\label{sec:introduction}
\begin{figure}[tb]
    \centering
	\includegraphics[width=0.85\textwidth]{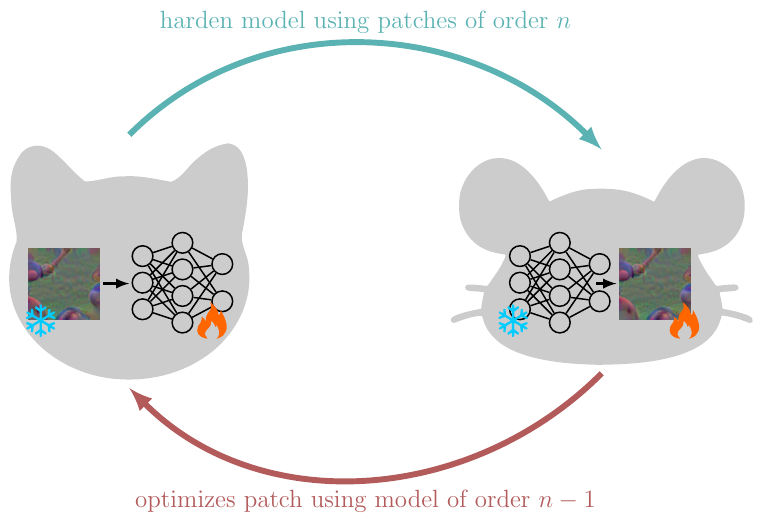}
	\caption{Playing the cat-and-mouse game: The mouse (attacker) tries to
	hide from the cat (defender). By integrating hardened models and
	optimized attack patterns into the respective optimization processes,
	a cat-and-mouse dynamic emerges.}
    \label{fig:higher-order-circle}
\end{figure}

Recently, machine learning has seen some remarkable advancements in various
applications. However, the vulnerability to adversarial attacks, where carefully
crafted inputs are used to deceive a model into making erroneous decisions,
remains. Due to their potential in exploiting and compromising the security and
reliability of deep neural networks, a critical area of studying these
phenomena has emerged. In the context of physical world attacks against object
detectors, adversarial patches are one of the simplest and most direct approaches
to fool these systems. Securing the models against these kinds of attacks can be
achieved naively by introducing examples of adversarial patterns during
the training. This so-called adversarial training results in a hardened model,
which is more secure against a 1st-order
attack~\cite{Goodfellow2015,madry2018towards}. When optimizing a patch on this
hardened 1st-order network, a 2nd-order patch can be optimized. This iterative
cat-and-mouse dynamic may be conducted multiple times (see
\autoref{fig:higher-order-circle}), potentially yielding more robust models and
enhanced patches.  However, the question arises: To what extent does repeated
engagement in this dynamic contribute to the robustness of the models and the
strength of the patches?

This work (i) investigates the behavior of both the patches and models when
playing the cat-and-mouse game with different parameterizations. (ii) The impact
of higher-order patches is analyzed, and the performance of higher-order hardened
networks is compared in an extensive evaluation. Moreover, (iii) the network
and dataset transferability of the higher-order patches is examined.
As a reference object detector, the YOLOv10b object detector is chosen. The
patches are optimized in an evasion attack manner.

The rest of this paper is organized as follows:
\Cref{sec:related_work} provides related work regarding adversarial attacks and
adversarial training. \Cref{sec:higher_order_patches} introduces our
methodology for analyzing the patches. The experimental setup is given in
\autoref{sec:experimental_setup} and results of the evaluation are presented in
\autoref{sec:evaluation}. A conclusion and a summary of our findings and
directions for future research are given in \autoref{sec:conclusion}.

\section{Related Work}
\label{sec:related_work}
As adversarial attacks in computer vision comprise several subtopics, the
following section only covers a fraction of works that are directly connected.
For a broader overview, the interested reader is referred to one of the
following surveys~\cite{Akhtar2018,Serban2020,Chakraborty2021,Wei2024}.
Multiple works use the term \emph{order} in relation to adversarial attacks,
yet most are about using higher-order derivative
information~\cite{Bai2018,Singla2020,Peng2024} or low-order and high-order
interactions~\cite{Ren2021}. \textbf{By the term order, we refer to the
number of cycles the adversarial counterplay dynamic has been played} (see
\autoref{fig:higher-order-circle}). Unfortunately, there are only a
few~\cite{cai2018curriculum,Lin2024,Zhang2022} other works that
investigate such a successive and iterative adversarial attack.

Curriculum adversarial training, as proposed by Cai
\etal~\cite{cai2018curriculum}, iteratively increases the attack strength until
an upper bound specified by the defender is reached. By doing so, they overcome
overfitting against attacks and significantly improve recent state-of-the-art
methods. They also state an attack generalization issue, where models trained
with weaker attacks may not generalize to stronger attacks.

Lin \etal~\cite{Lin2024} propose a successive perturbation generation scheme for
adversarial training (SPGAT) which successively strengthens adversarial
examples. Instead of generating adversarial examples from the original input in
each training epoch, they generate the adversarial examples based on previous
epochs.  Moreover, they shift models across the training epochs to enhance the
efficiency of adversarial training. By doing so, they harden an image
classifier against different attack schemes and conditions.

To boost the transferability of adversarial examples, Zhang
\etal~\cite{Zhang2022} propose successively attacking multiple high-accuracy
models and adding modest adversarial perturbations progressively across these
models to obtain adversarial examples targeting the standard vulnerable
directions of the models.

Contrary to the presented work, this study does not investigate adversarial
examples attacking image classifiers but rather adversarial
patches~\cite{Brown2017} attacking real-time object detectors. As these patches
can easily be used to attack object detectors in the physical
world~\cite{Thys2019,Wu2020,Wei2024,Li2025} and provide a high network and
dataset transferability~\cite{Bayer2024}, they pose a threat for every
non-hardened system.

\section{A Cat-and-Mouse Game}
\label{sec:higher_order_patches}
Wiktionary defines a cat-and-mouse game as a situation where two parties
closely monitor and challenge one another in a suspicious or self-protective
manner, often because each party is attempting to gain an advantage over the
other~\cite{wiktionary}. This basically describes the dynamic of an attacker
that successively optimizes his strategy while a defender reacts and hardens his
defense. To study this cat-and-mouse dynamic in real-time object detection,
first the attackers tactic (adversarial patches) and then the defenders response
(adversarial training) are analyzed.

\subsection{Adversarial Patches}
When it comes to physical adversarial attacks against object detectors,
adversarial patches with the goal to suppress a detection are one of the most
practically feasible attacks~\cite{Thys2019,Li2025,Wei2024}. In the investigated
setting, the attackers objective is to reduce the detectors confidence for the
selected class. Regarding the threat model, a white box scenario is
investigated. For the main experiments, the attacker has full access to the 
models. The transferability study can be considered a gray box scenario, as no
information about the target models besides the common training dataset is
considered during patch optimization.

To maintain realistic optimization conditions, the patch optimization process
applies an augmentation pipeline (e.g., scale, rotation, viewpoint) before
placing the patch inside of bounding boxes of objects of interest. The patches
are optimized by freezing the detector weights, propagating altered images of a
training dataset through the detector, and using the \emph{objectness} score of
instances of the target class as a loss term. In addition, a \emph{smoothness}
and \emph{validity} loss is used to reduce high-frequency information and
prevent \enquote{illegal} colors. The final optimization loss
\begin{equation}
	L = \lambda_\text{obj} L_\text{obj} + \lambda_\text{smt} L_\text{smt} +
	\lambda_\text{val} L_\text{val} 
\end{equation}
is the weighted sum of the objectness, smoothness, and validity loss.

During evaluation, there is a probability of $p_\text{box}$ that patches are
applied to instances of the investigated object class and of $p_\text{hal}$ that
they are hallucinated at a random position in the image to distract the
detector and punish overfitting. The patches are resized according to the
shorter side of the respective bounding box and placed in the center of it.

\subsection{Adversarial Training}
The defenders response in the cat-and-mouse game is adversarial training: the
detector is optimized to perform well even under worst-case patch overlays.
Patch-aware training is used in such a way that each ground-truth bounding box
instance of objects of interest in the training dataset has a certain
probability $\pi$ to contain an adversarial patch at a random position. The
probability is a crucial parameter, as a value that is too high would lead to a
detector, that assumes that the patch is a crucial part of the object of
interest, causing a higher false positive rate. If the probability is too low,
the desired effect of the robustness of the system against these patterns is not
given.

\subsection{Higher Order Patches}
Playing the cat-and-mouse game results in higher-order adversarial patches,
which are the evolution in attack strategies designed to counteract defenses
introduced by adversarial training. While traditional (1st-order) adversarial
patches are optimized to deceive a standard (0th-order) object detector,
(n+1)-th-order adversarial patches are crafted to bypass (n-th-order)
detectors that have been hardened against such attacks through adversarial
training.

The optimization process for higher-order patches involves two sequential
optimization processes: while the defenders model is adversarially trained
using current estimates of the adversarial patch, the attacker optimizes
patches to maximize its impact on the newly hardened model (see
\autoref{fig:higher-order-circle}).

\section{Experimental Setup}
\label{sec:experimental_setup}

\subsection{Conducted Experiments}
There are two key factors in adversarial training that are investigated in the
following experiments: the number of patches used and whether to include patches
from all previous orders during training. These factors are combined and result
in four distinct experimental settings:
\begin{itemize}
\item[]\textbf{k=1 patch optimized, non-successive training:} The n-th order
	model is exposed to the latest single n-th order patch. A single new
		patch (k=1) is optimized each order.
\item[]\textbf{k=1 patch optimized, successive training:} The n-th order
	model is exposed to all ($\leq n$)-th order patches. A single new patch
		(k=1) is optimized each order.
\item[]\textbf{k=3 patches optimized, non-successive training:} The
	n-th order model is exposed to all n-th order patches. Multiple new
		patches (k=3) are optimized each order.
\item[]\textbf{k=3 patches optimized, successive training:} The
	n-th order model is exposed to all ($\leq n$)-th order patches. Multiple
		new patches (k=3) are optimized each order.
\end{itemize}

Additionally, the optimized patches of these four experiments are evaluated in a
transferability study, where they are used to attack other detectors of the YOLO
family.

\subsection{Datasets}
The class of objects of interest for the experiments is \emph{Person}. To train
the object detectors, COCO~\cite{Lin2014} is used, as it is an established
benchmark dataset and most detectors are pre-trained on it. Furthermore, by far
the largest number of objects in COCO are persons; thus, the dataset provides a
good foundation. As COCO test-dev does not provide ground-truth annotations,
COCO \emph{val2017} is used as an evaluation split. Throughout the rest of the
paper, the evaluation set refers to \emph{val2017}.

In addition to COCO, a much smaller person-centered dataset is used: \emph{INRIA
Person}~\cite{Dalal2005}. The dataset consists of images depicting persons in
different environments. To optimize adversarial patches, the train split of the
positive image set is used.

\subsection{Object Detectors}
If not stated otherwise, all experiments use the YOLOv10b object detector
provided by Ultralytics~\cite{YOLO2023}. To enable adversarial training, the
augmentation pipeline has been modified slightly: after loading an image, a
random patch of the set of available patches is selected and applied inside
bounding boxes of interest.
To enable the patch optimization, some minor mandatory changes during the
inference phase have also been made. Moreover, as YOLOv10 no longer has an
objectness score, the pre-sigmoid prediction confidences of objects of interest
of both detector heads (one-to-one and one-to-many) are used instead.

\subsection{Adversarial Patch Optimization}
All optimized patches have a resolution of $256\times256$ px and are initialized
with random values in $[0,1]^3$. They are optimized in an evasion attack manner
over 150 epochs using AdamW with an initial learning rate of 0.01 and a
reduction of the learning rate every 50 epochs by a factor of 10.  The
augmentation pipeline applied to the patches consists of a random color jitter,
followed by a random rotation in $[-30^\circ,30^\circ]$ and a random perspective
transformation. The augmented patch is then randomly resized to match the
smaller size of the corresponding bounding box by a factor of $[0.3,0.6]$. The
patches are optimized using \emph{INRIAPerson} as a training dataset and use
\emph{Person} as the target class.

\subsection{Adversarial Training}
The adversarial training procedure is based on the regular training script
provided in the \emph{Ultralytics} repository~\cite{YOLO2023}. An additional
augmentation step after the images are loaded is added: the images are
propagated through a patch applier that modifies an image by placing an
adversarial patch in the image at a random position inside the ground-truth
bounding box with a probability of $\pi=0.25$. Unlike during patch training, the
patches are not augmented but rather only resized randomly to match the smaller
size of the corresponding bounding box by a factor of $[0.75,0.9]$.
The patch application happens before mosaicing and any other augmentations.
Each model is optimized with AdamW over 100 epochs using the COCO training
split and the default hyperparameters presented by Ultralytics.

\section{Evaluation}
\label{sec:evaluation}
The performance of the detectors is measured using the mean average precision
(mAP) with multiple intersection over union (IoU) thresholds~\cite{Lin2014}.
Since we focus solely on the \emph{Person} class, the class mean collapses to
the class average precision; consequently, the AP$_{\text{Person@[.5:.95]}}$ (AP)
is reported.

With a probability of $p_{box}=0.5$, an optimized patch is placed at the center
of ground-truth bounding boxes in images of the evaluation set. Additionally,
there is a chance of $p_{hal}=0.5$ that the same patch is hallucinated and
placed at random positions in the image. To ensure a reproducible evaluation, a
fixed random seed is used. This way, images of different evaluations only differ
in the content of applied patches.
The patches are resized to match 50\% of the shorter side of the respective
bounding box. This way, the patches have a strong effect without completely
covering a \emph{Person} instance. In addition to the 50\% resize factor, the
appendix also covers results for resize factors of 25\% and 75\%.

\subsection{Baseline}
The first step is to produce reference points for the object detectors. Towards
this, the performance of each trained detector is investigated on the evaluation
set when no patches are present. As presented in \autoref{tbl:baseline} the
detector performs slightly worse when grayscale patches are applied compared to
the clean evaluation set. This is expected behavior, as the introduced
occlusion of objects usually leads to a small performance drop. 
\begin{table}[tb]
\centering

\begin{tabular}{cccrr}
\toprule
	Successive & k & n & AP$_\text{clean}$ & AP$_\text{Grayscale}$\\
	\midrule
	\xmark & 1 & 0 & 0.59 & 0.52\\
	\xmark & 1 & 1 & 0.58 & 0.52\\
	\xmark & 1 & 2 & 0.58 & 0.52\\
	\xmark & 1 & 3 & 0.58 & 0.52\\
	\xmark & 1 & 4 & 0.58 & 0.52\\
	\xmark & 1 & 5 & 0.59 & 0.52\\
\bottomrule
\end{tabular}\hfill
\begin{tabular}{cccrr}
\toprule
	Successive & k & n & AP$_\text{clean}$ & AP$_\text{Grayscale}$\\
	\midrule
	\xmark & 3 & 0 & 0.59 & 0.52\\
	\xmark & 3 & 1 & 0.58 & 0.54\\
	\xmark & 3 & 2 & 0.58 & 0.53\\
	\xmark & 3 & 3 & 0.58 & 0.53\\
	\xmark & 3 & 4 & 0.58 & 0.54\\
	\xmark & 3 & 5 & 0.58 & 0.53\\
\bottomrule
\end{tabular}

\begin{tabular}{cccrr}
\toprule
	Successive & k & n & AP$_\text{clean}$ & AP$_\text{Grayscale}$\\
	\midrule
	\cmark & 1 & 0 & 0.59 & 0.52\\
	\cmark & 1 & 1 & 0.58 & 0.54\\
	\cmark & 1 & 2 & 0.58 & 0.54\\
	\cmark & 1 & 3 & 0.58 & 0.54\\
	\cmark & 1 & 4 & 0.58 & 0.54\\
	\cmark & 1 & 5 & 0.58 & 0.54\\
\bottomrule
\end{tabular}\hfill
\begin{tabular}{cccrr}
\toprule
	Successive & k & n & AP$_\text{clean}$ & AP$_\text{Grayscale}$\\
	\midrule
	\cmark & 3 & 0 & 0.59 & 0.52\\
	\cmark & 3 & 1 & 0.58 & 0.54\\
	\cmark & 3 & 2 & 0.58 & 0.54\\
	\cmark & 3 & 3 & 0.58 & 0.54\\
	\cmark & 3 & 4 & 0.58 & 0.54\\
	\cmark & 3 & 5 & 0.58 & 0.54\\
\bottomrule
\end{tabular}

	\caption{Baseline detector performance where AP$_\text{clean}$ is the
	performance of the detector on the evaluation set when no patches are
	present and AP$_\text{Grayscale}$ when 11 different grayscale levels are
	used instead of patches. The standard deviations for
	AP$_\text{Grayscale}$ are omitted, as they are below 0.01.}
\label{tbl:baseline}
\end{table}

\subsection{Cat-and-Mouse-Game}
\begin{figure}[tb]
	\centering
	\includegraphics[width=0.45\textwidth]{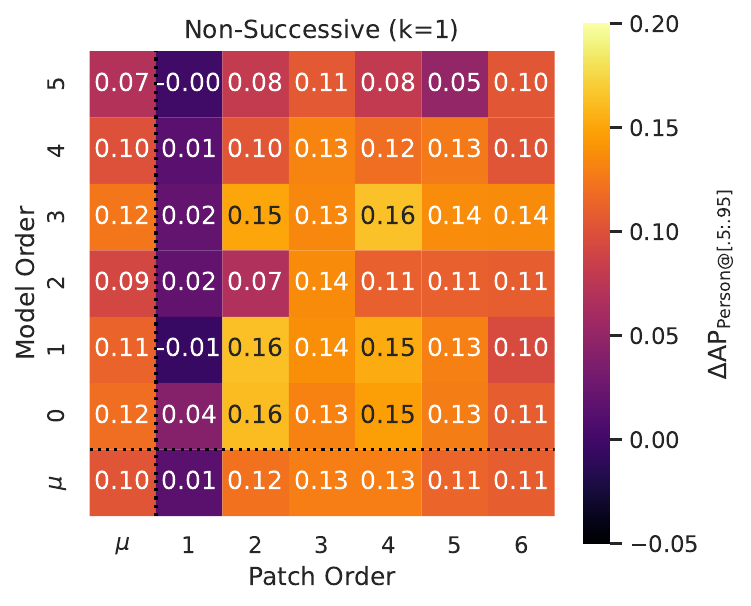}
	\hfill
	\includegraphics[width=0.45\textwidth]{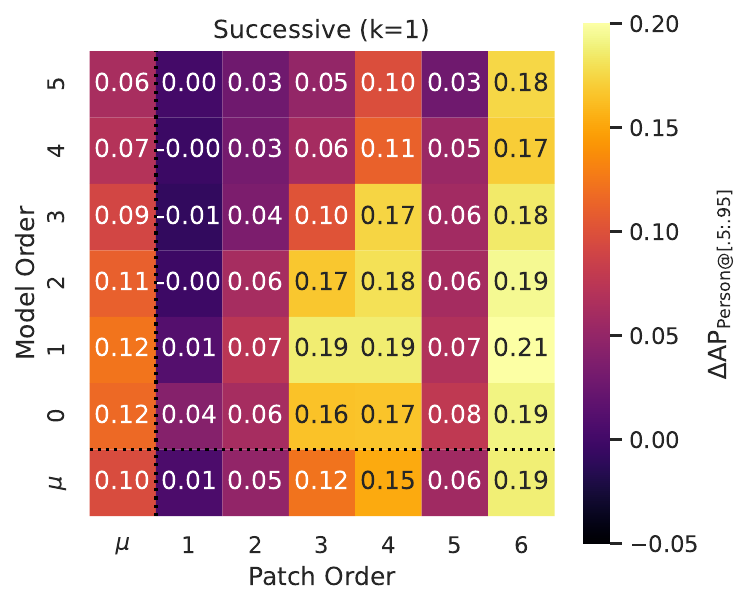}
	\hfill
	\includegraphics[width=0.45\textwidth]{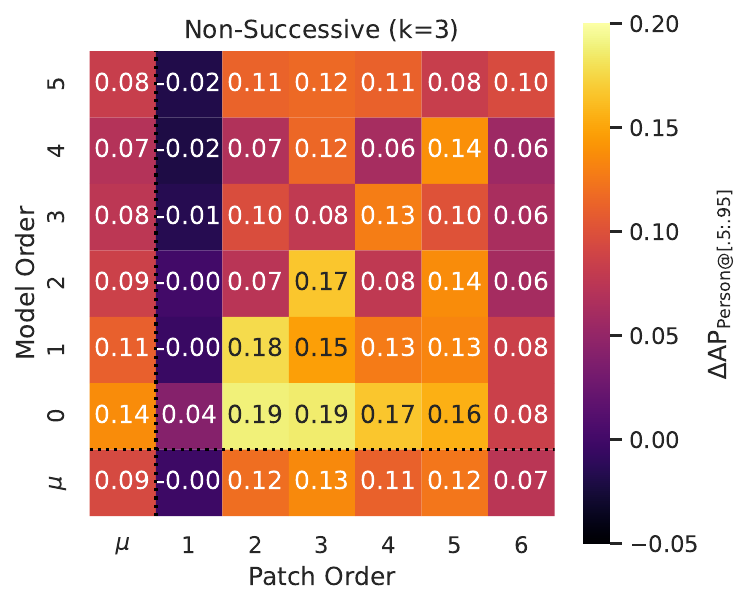}
	\hfill
	\includegraphics[width=0.45\textwidth]{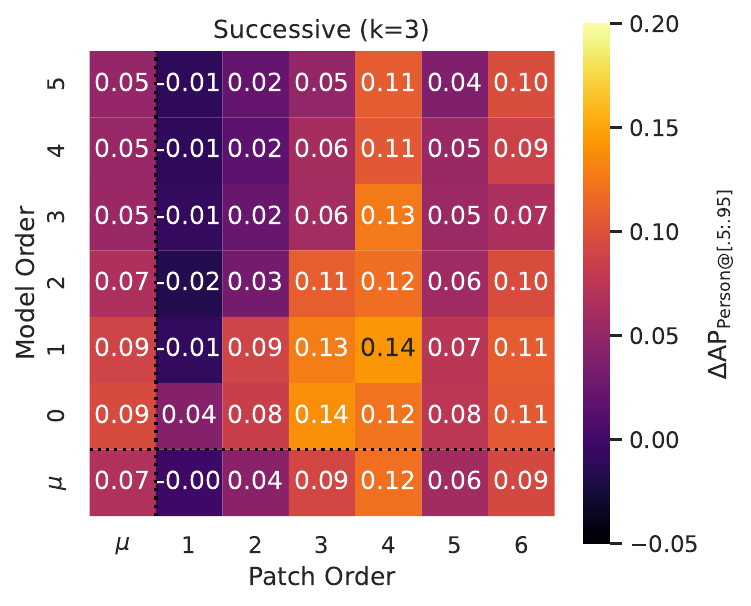}
	\caption{Heatmap of the performance of higher-order patches and
	higher-order networks on the evaluation set. The columns represent the
	order of the tested adversarial patches, while the rows indicate the
	order of the evaluated network. The shade of each cell encodes the
	arithmetic mean of the achieved $\Delta\text{AP}=AP_\text{Grayscale} -
	AP$ of a network for a set of 4 validation patches. $\mu$ is the row-
	and columnwise arithmetic mean.}
	\label{fig:heatmap}
\end{figure}

The results of the main experiment are given in \autoref{fig:heatmap}. 
Each subplot corresponds to one experiment. \emph{Non-Successive~(k=1)} and
\emph{Non-Successive~(k=3)} are the experiments where the lower-order patches
are not included in adversarial training. \emph{Successive~(k=1)} and
\emph{Successive~(k=3)}, on the other hand, include all previous lower-order
patches in the adversarial training.

Each column of a heatmap corresponds to a different order of tested adversarial
patches, while each row is related to the order of the evaluated network.
The shade of each cell encodes the achieved AP of a network for a set
of 4 validation patches. These patches are optimized with the same parameters
utilizing the same detectors as the corresponding train patches but are not
included during adversarial training. The separated rows and columns labeled
$\mu$ are the arithmetic mean of the corresponding columns and rows.

\emph{Non-Successive~(k=1)} is the simplest case where only a single patch is
used for adversarial training. The resulting plot shows a comparable bad
performance regardless of the patch order or model order. The highest $\Delta$AP
is at 0.16, while the mean is at $0.10\pm 0.06$. As expected, all adversarial
trained models perform better than the regular trained model when attacked with
a 1st-order patch. Moreover, higher-order patches have on average a much
higher impact on the detector performances than 1st-order patches.

The benefit of including additional patches during adversarial training is
presented in experiment \emph{Non-Successive~(k=3)}. The performance of
higher-order models when attacked with higher-order patches improves. In
addition, the main diagonal becomes visible, indicating that the adversarial
training improves the performance of the networks when attacking with lower
adversarial patches. Yet, this is only true for networks with an order equal to
one plus the current patch order. For patches of order 1, 2, 3, and 4, models
of order 2, 3, 4, and 5 perform worse than models of order 1, 2, 3, and 4. This
suggests that there is no sufficient patch diversity present.

Consequently, the \emph{Successive~(k=1)} and \emph{Successive~(k=3)}
experiments are conducted, where all previously optimized patches are included
in the current adversarial training. The results for \emph{Successive~(k=1)} and
\emph{Successive~(k=3)} are quite similar: As expected, higher-order networks
profit from including previous patches. The observed insufficient patch
diversity as in \emph{Non-Successive~(k=3)} is no longer present.
Interestingly, the overall performance of the networks in
\emph{Successive~(k=3)} is higher compared to \emph{Successive~(k=1)} even for
the lower-order networks, which could be due to either more robust networks or
less impactful patches. The conducted transferability study gives a hint on how
to interpret this observation.

Depending on the experimental setup, a clear trend or prediction on how even
higher-order patches would perform is hardly possible. However, a cautious
prediction can be made: there is a certain trend that higher-order models,
regardless of the setup, have a higher AP, averaged over the different patch
orders. Higher-order patches in the \emph{Non-Successive} cases do not seem to
lower the AP on average at all, while in the \emph{Successive} cases, a
\enquote{wavy} downward trend becomes visible.

\subsection{Transferability}
\begin{figure}[tb]
	\centering
	\includegraphics[width=0.45\textwidth]{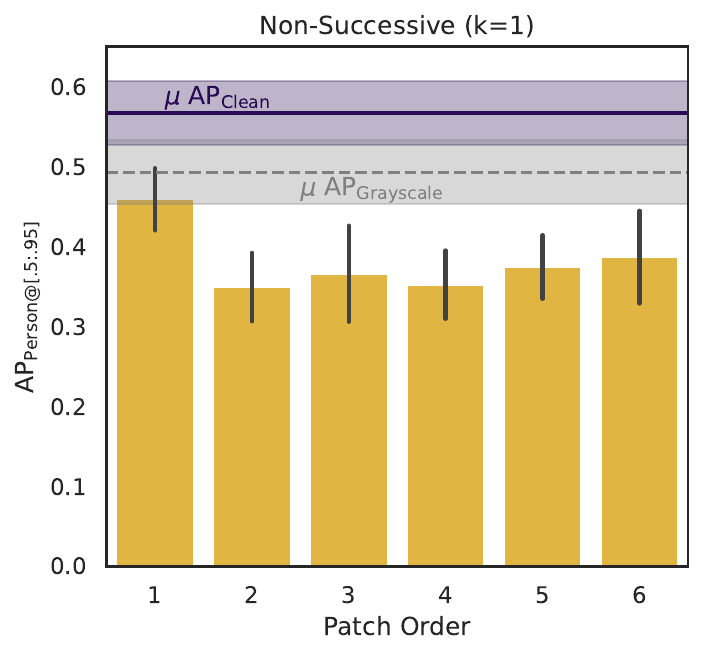}
	\hfill
	\includegraphics[width=0.45\textwidth]{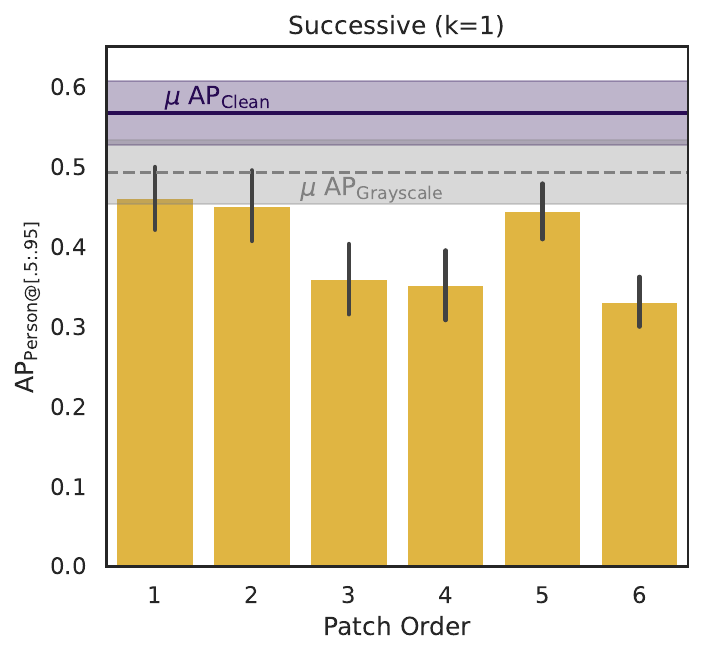}
	\hfill
	\includegraphics[width=0.45\textwidth]{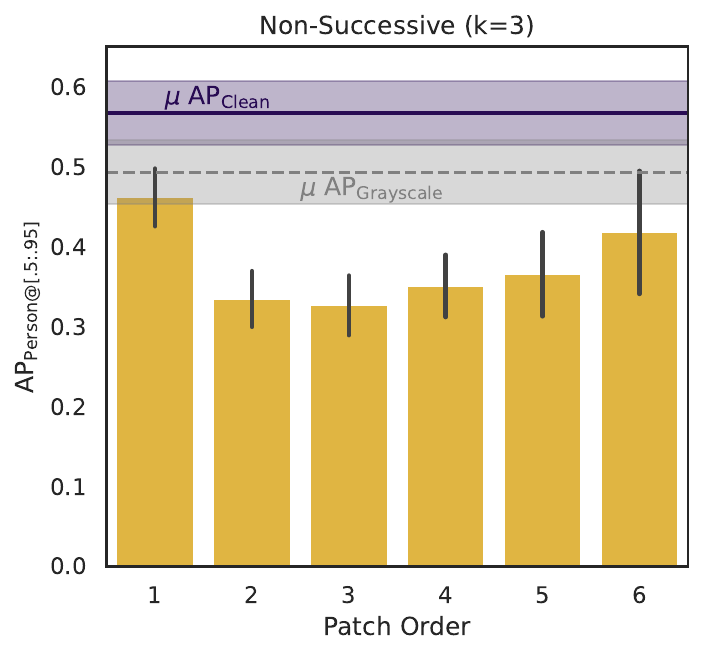}
	\hfill
	\includegraphics[width=0.45\textwidth]{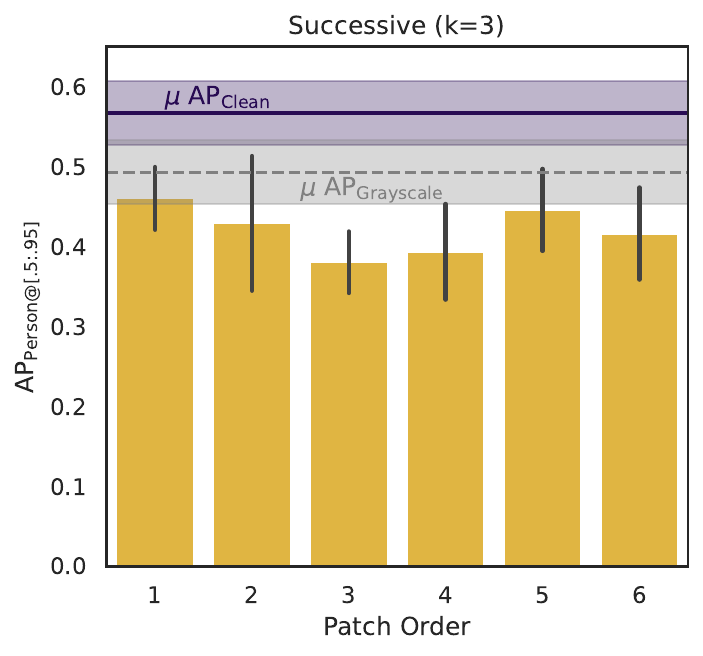}
	\caption{Results of the transferability study. Each bar shows the mean
	performance of 21 detectors for the corresponding patches of the given
	order. In addition, the standard deviation is given.}
	\label{fig:transferability}
\end{figure}

Regarding the transferability of the optimized higher-order patches,
\autoref{fig:transferability} shows the aggregated performance of the
investigated 21 COCO-pretrained object detectors
(YOLOv9/10/11/12)~\cite{wang2024yolov9,wang2024yolov10,yolo11_ultralytics,tian2025yolov12}.
Each bar shows the mean performance for the corresponding validation patches of
the given order. In addition, the standard deviation for each bar and the
average clean and grayscale performance are given.

All investigated models are of order 0 and are thus not hardened against
adversarial patches. Again, the mentioned \enquote{wavy} downward trend in the
\emph{Successive} cases becomes visible. In addition, the question whether the
patches in \emph{Successive(k=3)} become less impactful can be answered.
Compared to \emph{Successive(k=1)}, the AP for patches of order 3, 4, and 6 is
lower and thus reinforces the assumption of a decrease in patch performance.  A
similar observation can be made in the \emph{Non-Successive} cases: a slightly
upward trend for k=1 and a strong upward trend for k=3 become visible.  A
possible reason for this behavior could be that the internal activations of
lower-order models are more similar to the investigated 0th-order models.  Yet,
the impact of 2nd-order patches for the \emph{Non-Successive} case is
significantly higher than the 1st-order patches.

\subsection{Discussion}
While the presented results provide an insight into the performance of higher-order
patches and models, the results should be treated with caution. We do not cover
and consider the impact of the distribution shift caused by different
parameterization in model hardening and patch optimization, such as models
trained and evaluated with COCO while the patches are optimized using
\emph{INRIAPerson} or different resize ranges. A change in the patch
parameterization, such as the resize range, initial learning rate, number of
epochs, or the optimizer, could also lead to stronger patches. Moreover, the
experiments are solely conducted on the \emph{Person} class, while the detectors
are trained regularly using the complete 80 COCO classes. Nevertheless, the focus of the
presented study is not to find optimal parameters but rather to get a more
profound understanding of adversarial patches by an initial attempt at playing
the cat-and-mouse game.
% Verteilungsshift der resize range noch mit reinbringen

A problem we experienced with the presented large-scale evaluation of
adversarial patches is that the optimization of a single patch takes a
comparably large amount of time (about 3.5 h with an Nvidia RTX4090 and an
Intel i9-7980XE CPU). This is a result of a technical limitation in our current
optimization strategy, which does not seem to improve in the near future. As a
result, investigating even more higher-order adversarial patches and therefore
indicating a clear trend of the patch performance could only be achieved with a
reduced input dimensionality.

\section{Conclusion}
\label{sec:conclusion}
The presented work investigates higher-order adversarial patches for real-time
object detection. Higher-order patches are the result of successive optimizing
adversarial patches and harden the object detector in a cat-and-mouse game
manner. As a representative object detector, the extensive evaluation uses the
popular YOLOv10 detector. Four different experiments are conducted to
investigate two key factors: including previously used adversarial patches in
the current adversarial training and the impact of using multiple adversarial
patches compared to a single patch. The results indicate that higher-order
adversarial patches have a stronger impact on the detector performance than
1st-order adversarial patches. Moreover, higher-order detectors perform better
than lower-order ones. Yet, the performance of the hardened models when
attacked with even higher-order patches is only partly restored to the
grayscale-patched and clean performance of the networks, supporting the thesis of
an attack generalization issue in object detection similar to the one mentioned
by Cai \etal~\cite{cai2018curriculum} in image classification.

\subsection{Future Work}
Future work should investigate even more higher-order adversarial patches, check
if there is a clear trend, and try to answer the question of whether the
cat-and-mouse game converges or diverges. If there is an indication of
accumulation points, then these should be used to enhance the hardening of
object detectors and perhaps neural networks in general against adversarial
attacks.

\bibliographystyle{unsrtnat}
\bibliography{bib}

@String(CVPR= {IEEE Conf. Comput. Vis. Pattern Recog.})

@String(NIPS= {Adv. Neural Inform. Process. Syst.})

@String(ICLR = {Int. Conf. Learn. Represent.})

@String(CVPR  = {CVPR})

@String(NIPS  = {NeurIPS})

@String(ICLR  = {ICLR})

@inproceedings{
madry2018towards,
title={Towards Deep Learning Models Resistant to Adversarial Attacks},
author={Aleksander Madry and Aleksandar Makelov and Ludwig Schmidt and Dimitris Tsipras and Adrian Vladu},
booktitle={International Conference on Learning Representations},
year={2018},
url={https://openreview.net/forum?id=rJzIBfZAb},
}

@inproceedings{cai2018curriculum,
  title={Curriculum Adversarial Training},
  author={Cai, Qi-Zhi and Liu, Chang and Song, Dawn},
  booktitle={Proceedings of the Twenty-Seventh International Joint Conference on Artificial Intelligence},
  pages={3740--3747},
  year={2018},
  organization={International Joint Conferences on Artificial Intelligence Organization}
}

@article{tian2025yolov12,
  title={Yolov12: Attention-centric real-time object detectors},
  author={Tian, Yunjie and Ye, Qixiang and Doermann, David},
  journal={arXiv preprint arXiv:2502.12524},
  year={2025}
}

@software{yolo11_ultralytics,
  author = {Glenn Jocher and Jing Qiu},
  title = {Ultralytics YOLO11},
  version = {11.0.0},
  year = {2024},
  url = {https://github.com/ultralytics/ultralytics},
  orcid = {0000-0001-5950-6979, 0000-0003-3783-7069},
  license = {AGPL-3.0}
}

@article{wang2024yolov10,
  title={Yolov10: Real-time end-to-end object detection},
  author={Wang, Ao and Chen, Hui and Liu, Lihao and Chen, Kai and Lin, Zijia and Han, Jungong and others},
  journal={Advances in Neural Information Processing Systems},
  volume={37},
  pages={107984--108011},
  year={2024}
}

@inproceedings{wang2024yolov9,
  title={Yolov9: Learning what you want to learn using programmable gradient information},
  author={Wang, Chien-Yao and Yeh, I-Hau and Mark Liao, Hong-Yuan},
  booktitle={European conference on computer vision},
  pages={1--21},
  year={2024},
  organization={Springer}
}

@inproceedings{Li2025,
author = {Li, Yanjie and Liang, Kaisheng and Xiao, Bin},
booktitle = {ICLR 2025},
title = {{UV-Attack: Physical-World Adversarial Attacks on Person Detection via Dynamic-NeRF-based UV Mapping}},
year = {2025}
}

@article{Goodfellow2015,
archivePrefix = {arXiv},
arxivId = {1412.6572},
author = {Goodfellow, Ian J. and Shlens, Jonathon and Szegedy, Christian},
eprint = {1412.6572},
journal = {3rd International Conference on Learning Representations, ICLR 2015 - Conference Track Proceedings},
pages = {1--11},
title = {{Explaining and harnessing adversarial examples}},
year = {2015}
}

@misc{wiktionary,
   author = "Wiktionary",
   title = "cat and mouse --- Wiktionary{,} The Free Dictionary",
   year = "2025",
   url = "{https://en.wiktionary.org/w/index.php?title=cat\_and\_mouse\&oldid=85469831}",
   note = "[Online; accessed 2-December-2025]"
 }

@article{Wei2024,
archivePrefix = {arXiv},
arxivId = {2209.15179},
author = {Wei, Hui and Tang, Hao and Jia, Xuemei and Wang, Zhixiang and Yu, Hanxun and Li, Zhubo and Satoh, Shinichi and {Van Gool}, Luc and Wang, Zheng},
doi = {10.1109/TPAMI.2024.3430860},
eprint = {2209.15179},
issn = {19393539},
journal = {IEEE Trans. Pattern Anal. Mach. Intell.},
number = {12},
pages = {9797--9817},
publisher = {IEEE},
title = {{Physical Adversarial Attack Meets Computer Vision: A Decade Survey}},
volume = {46},
year = {2024}
}

@article{Wu2020,
archivePrefix = {arXiv},
arxivId = {1910.14667},
author = {Wu, Zuxuan and Lim, Ser Nam and Davis, Larry S. and Goldstein, Tom},
doi = {10.1007/978-3-030-58548-8_1},
eprint = {1910.14667},
isbn = {9783030585471},
issn = {16113349},
journal = {Lect. Notes Comput. Sci.},
pages = {1--17},
title = {{Making an Invisibility Cloak: Real World Adversarial Attacks on Object Detectors}},
volume = {12349 LNCS},
year = {2020}
}

@inproceedings{Brown2017,
  title={Adversarial patch},
  author={Brown, {Tom B} and Mane, Dandelion and Aurko, Roy and Abadi, Martin and Gilmer, Justin},
  booktitle={31st Conference on Neural Information Processing Systems (NIPS 2017)},
  year={2017}
}

@article{Zhang2022,
author = {Zhang, Xiaolin and Zhang, Wenwen and Liu, Lixin and Wang, Yongping and Gao, Lu and Zhang, Shuai},
doi = {10.2139/ssrn.4162643},
journal = {SSRN Electron. J.},
number = {2},
pages = {306--316},
title = {{Enhancing Transferability of Adversarial Examples by Successively Attacking Multiple Models}},
volume = {25},
year = {2022}
}

@article{Lin2024,
author = {Lin, Wei and Liao, Lichuan},
doi = {10.1631/FITEE.2300474},
issn = {20959230},
journal = {Front. Inf. Technol. Electron. Eng.},
number = {4},
pages = {527--539},
title = {{Towards sustainable adversarial training with successive perturbation generation}},
volume = {25},
year = {2024}
}

@article{Chakraborty2021,
author = {Chakraborty, Anirban and Alam, Manaar and Dey, Vishal and Chattopadhyay, Anupam and Mukhopadhyay, Debdeep},
doi = {10.1049/cit2.12028},
issn = {24682322},
journal = {CAAI Trans. Intell. Technol.},
number = {1},
pages = {25--45},
title = {{A survey on adversarial attacks and defences}},
volume = {6},
year = {2021}
}

@article{Serban2020,
archivePrefix = {arXiv},
arxivId = {2008.04094},
author = {Serban, Alex and Poll, Erik and Visser, Joost},
doi = {10.1145/3398394},
eprint = {2008.04094},
issn = {15577341},
journal = {ACM Comput. Surv.},
number = {3},
title = {{Adversarial Examples on Object Recognition: A Comprehensive Survey}},
volume = {53},
year = {2020}
}

@inproceedings{Dalal2005,
author = {Dalal, N. and Triggs, B.},
booktitle = {CVPR},
doi = {10.1109/CVPR.2005.177},
isbn = {0-7695-2372-2},
issn = {2313433X},
pages = {886--893},
publisher = {IEEE},
title = {{Histograms of Oriented Gradients for Human Detection}},
volume = {1},
year = {2005}
}

@article{Lin2014,
archivePrefix = {arXiv},
arxivId = {1405.0312},
author = {Lin, Tsung Yi and Maire, Michael and Belongie, Serge and Hays, James and Perona, Pietro and Ramanan, Deva and Doll{\'{a}}r, Piotr and Zitnick, C. Lawrence},
doi = {10.1007/978-3-319-10602-1_48},
eprint = {1405.0312},
issn = {16113349},
journal = {Lecture Notes in Computer Science (including subseries Lecture Notes in Artificial Intelligence and Lecture Notes in Bioinformatics)},
number = {PART 5},
pages = {740--755},
title = {{Microsoft COCO: Common objects in context}},
volume = {8693 LNCS},
year = {2014}
}

@software{YOLO2023,
author = {Jocher, Glenn and Qiu, Jing and Chaurasia, Ayush},
license = {AGPL-3.0},
month = jan,
title = {{Ultralytics YOLO}},
url = {https://github.com/ultralytics/ultralytics},
version = {8.0.0},
year = {2023}
}

@article{Ren2021,
archivePrefix = {arXiv},
arxivId = {2111.03536},
author = {Ren, Jie and Zhang, Die and Wang, Yisen and Chen, Lu and Zhou, Zhanpeng and Chen, Yiting and Cheng, Xu and Wang, Xin and Zhou, Meng and Shi, Jie and Zhang, Quanshi},
eprint = {2111.03536},
isbn = {9781713845393},
issn = {10495258},
journal = {Advances in Neural Information Processing Systems},
number = {NeurIPS},
pages = {3797--3810},
title = {{A Unified Game-Theoretic Interpretation of Adversarial Robustness}},
volume = {5},
year = {2021}
}

@article{Akhtar2018,
archivePrefix = {arXiv},
arxivId = {1801.00553},
author = {Akhtar, Naveed and Mian, Ajmal},
doi = {10.1109/ACCESS.2018.2807385},
eprint = {1801.00553},
issn = {2169-3536},
journal = {IEEE Access},
pages = {14410--14430},
publisher = {IEEE},
title = {{Threat of Adversarial Attacks on Deep Learning in Computer Vision: A Survey}},
volume = {6},
year = {2018}
}

@article{Thys2019,
archivePrefix = {arXiv},
arxivId = {1904.08653},
author = {Thys, Simen and Ranst, Wiebe Van and Goedeme, Toon},
doi = {10.1109/CVPRW.2019.00012},
eprint = {1904.08653},
isbn = {9781728125060},
issn = {21607516},
journal = {CVPR Workshops},
pages = {49--55},
title = {{Fooling automated surveillance cameras: Adversarial patches to attack person detection}},
volume = {2019-June},
year = {2019}
}

@article{Peng2024,
author = {Peng, Anjie and Shi, Guoqiang and Lin, Zhi and Zeng, Hui and Yang, Xing},
doi = {10.26599/TST.2024.9010154},
journal = {Tsinghua Sci. Technol.},
title = {{Approximating High-order Adversarial Attacks Using Runge-Kutta Methods}},
year = {2024}
}

@article{Bai2018,
archivePrefix = {arXiv},
arxivId = {arXiv:1809.03113v1},
author = {Bai, Li and Changyou, Chen and Wang, Wenlin and Carin, Lawrence},
eprint = {arXiv:1809.03113v1},
number = {2017},
pages = {1--14},
title = {{Second-Order Adversarial Attack and Certifiable Robustness}},
year = {2018}
}

@article{Singla2020,
archivePrefix = {arXiv},
arxivId = {2006.00731},
author = {Singla, Sahil and Feizi, Soheil},
eprint = {2006.00731},
isbn = {9781713821120},
journal = {37th International Conference on Machine Learning, ICML 2020},
pages = {8928--8938},
title = {{Second-order provable defenses against adversarial attacks}},
volume = {PartF16814},
year = {2020}
}

@inproceedings{Bayer2024,
author = {Jens Bayer and Stefan Becker and David M{\"u}nch and Michael Arens},
title = {{Network transferability of adversarial patches in real-time object detection}},
volume = {13206},
booktitle = {Artificial Intelligence for Security and Defence Applications II},
publisher = {SPIE},
pages = {132060X},
year = {2024},
doi = {10.1117/12.3031501},
URL = {https://doi.org/10.1117/12.3031501}
}

\end{document}